\documentclass[letterpaper, 10 pt, journal, twoside]{IEEEtran}
\usepackage{cite}
\ifCLASSINFOpdf
\else
\fi
\usepackage{amsmath}
\usepackage{array}
\usepackage{overpic}
\usepackage{amssymb}
\usepackage{multirow}
\usepackage{threeparttable}
\usepackage{hyperref}
\usepackage{fancyhdr}

\begin{document}
\title{Exploring the Mutual Influence between Self-Supervised Single-Frame and Multi-Frame Depth Estimation}

% \author{Michael~Shell,~\IEEEmembership{Member,~IEEE,}
%         John~Doe,~\IEEEmembership{Fellow,~OSA,}
%         and~Jane~Doe,~\IEEEmembership{Life~Fellow,~IEEE}% <-this % stops a space
% \thanks{M. Shell was with the Department
% of Electrical and Computer Engineering, Georgia Institute of Technology, Atlanta,
% GA, 30332 USA e-mail: (see http://www.michaelshell.org/contact.html).}% <-this % stops a space
% \thanks{J. Doe and J. Doe are with Anonymous University.}% <-this % stops a space
% \thanks{Manuscript received April 19, 2005; revised August 26, 2015.}}
\author{Jie Xiang$^{1, 2}$, Yun Wang$^{1}$, Lifeng An$^{1}$, Haiyang Liu$^{1}$ and Jian Liu$^{1}$%
\thanks{Manuscript received: April, 25, 2023; Revised July, 17, 2023; Accepted August, 17, 2023.}%Use only for final RAL version
\thanks{This paper was recommended for publication by Editor Cesar Cadena Lerma upon evaluation of the Associate Editor and Reviewers' comments.
This work was supported by the National Key Research and Development Program of China under Grant 2021YFB2501403.(Corresponding author: Yun Wang.)} %Use only for final RAL version
\thanks{$^{1}$Jie Xiang, Yun Wang, Lifeng An, Haiyang Liu and Jian Liu are with the Institute of Microelectronics of the Chinese Academy of Sciences, Beijing 100029, China%Jie Xiang, Yun Wang, Lifeng An, Haiyang Liu and Jian Liu
        {\tt\footnotesize (Email: \{xiangjie, wangyun, anlifeng, liuhaiyang, liujian\}@ime.ac.cn)}}%
\thanks{$^{2} $Jie Xiang is also with the University of  Chinese Academy of Sciences, Beijing 100049, China
        }%
\thanks{Digital Object Identifier (DOI): see top of this page.}
}
\maketitle
\thispagestyle{fancy}
\lhead{\small\copyright 2023 IEEE.  Personal use of this material is permitted.  Permission from IEEE must be obtained for all other uses, in any current or future media, including reprinting/republishing this material for advertising or promotional purposes, creating new collective works, for resale or redistribution to servers or lists, or reuse of any copyrighted component of this work in other works.}
\begin{abstract}
Although both self-supervised single-frame and multi-frame depth estimation methods only require unlabeled monocular videos for training, the information they leverage varies because single-frame methods mainly rely on appearance-based features while multi-frame methods focus on geometric cues. Considering the complementary information of single-frame and multi-frame methods, some works attempt to leverage single-frame depth to improve multi-frame depth. However, these methods can neither exploit the difference between single-frame depth and multi-frame depth to improve multi-frame depth nor leverage multi-frame depth to optimize single-frame depth models. To fully utilize the mutual influence between single-frame and multi-frame methods, we propose a novel self-supervised training framework. Specifically, we first introduce a pixel-wise adaptive depth sampling module guided by single-frame depth to train the multi-frame model. Then, we leverage the minimum reprojection based distillation loss to transfer the knowledge from the multi-frame depth network to the single-frame network to improve single-frame depth. Finally, we regard the improved single-frame depth as a prior to further boost the performance of multi-frame depth estimation. Experimental results on the KITTI and Cityscapes datasets show that our method outperforms existing approaches in the self-supervised monocular setting. 
\end{abstract}

% \begin{IEEEkeywords}
% IEEE, IEEEtran, journal, \LaTeX, paper, template.
% \end{IEEEkeywords}

% \fi
\IEEEpeerreviewmaketitle

\section{Introduction}
% for IEEE journal papers produced under \LaTeX\ using
% IEEEtran.cls version 1.8b and later.
\IEEEPARstart{D}{epth} estimation is an essential and challenging problem in 3D vision, which can be applied in a wide range of applications such as autonomous driving \cite{geiger2012we} and augmented reality \cite{luo2020consistent}. Although active depth sensors e.g. Lidar and binocular cameras-based methods \cite{guo2019group} exist, monocular depth estimation (MDE) methods that only use a single RGB camera to estimate depth still attract much attention due to their flexibility and low cost.

In the past years, many deep learning-based MDE methods \cite{eigen2014depth, garg2016unsupervised, zhou2017unsupervised, godard2019digging, Watson_2021_CVPR} have emerged. Among these methods, self-supervised methods \cite{zhou2017unsupervised, Watson_2021_CVPR} that use unlabeled monocular video sequences as training data to eliminate the dependence on ground-truth depth made exciting progress. Early self-supervised methods \cite{zhou2017unsupervised, godard2019digging, Watson_2021_CVPR} mainly focus on single-frame depth estimation, which refers to inferring the corresponding depth map given a single image. Whilst flexible, single-frame approaches ignore that more than one frame may be available at test time in many practical applications. Therefore, a few recent works \cite{Watson_2021_CVPR, feng2022disentangling, Wang2022CraftingMC, long_two} take multiple frames as input for depth estimation. Different from single-frame methods, these multi-frame methods mainly utilize the geometric matching features between multiple frames. Considering that dense feature matching is easily affected by occlusions, moving objects, and textureless regions, these methods try to utilize the single-frame information to improve multi-frame depth estimation unilaterally and show promising results.

However, these ``one-way'' methods do not take full advantage of the ``mutual'' influence between single-frame depth estimation and multi-frame depth estimation due to ignoring the two issues: (1) the potential benefits of the difference between single-frame depth and multi-frame depth, and (2) the effect of multi-frame depth on single-frame depth estimation. Since current multi-frame depth methods \cite{Wang2022CraftingMC, guizilini20203d} usually predict more accurate depth than single-frame methods \cite{godard2019digging, he2022ra}, the output of multi-frame depth models can be regarded as pseudo labels. Then we can model the uncertainty of single-frame depth using the difference between single-frame depth and multi-frame depth. Based on the above idea, we propose the Pixel-wise Adaptive Depth Sampling (PADS) module to determine the depth candidates used for multi-frame depth estimation, in which the single-frame depth is used to determine the geometric center of sampling range following \cite{Wang2022CraftingMC, long_two}, and the difference between single-frame depth and multi-frame depth is used to determine the width of the sampling range. In this way, we improve the efficiency of depth sampling and form effective cost volumes for more accurate multi-frame depth estimation. Regarding the second issue, we adopt the multi-frame depth model as the teacher to train another single-frame depth network via distillation learning. To alleviate the impact that the teacher model generates inaccurate labels, we combine the photometric loss \cite{godard2019digging} and the distillation loss to form a minimum reprojection based distillation loss, ignoring the pseudo labels with large reprojection errors. Thus, the single-frame depth network also produces better results.

To allow these two ideas to work in a compatible manner, we further propose a novel self-supervised training framework. Specifically, an uncertainty map is iteratively updated using the PADS module when training the teacher model and then fixed at test time. After distillation learning, the learned uncertainty map and the improved single-frame depth are further regarded as the input of the PADS module to determine the sampling range for cost volume generation so that the multi-frame network also benefits from the improved single-frame network.

In summary, the contributions of the paper are as follows:
\begin{itemize}
	\item A novel self-supervised distillation learning framework for MDE that fully utilizes the mutual benefits between self-supervised single-frame and multi-frame depth estimation.
	\item A pixel-wise adaptive depth sampling module to use the single-frame depth and the difference between single-frame depth and multi-frame depth as priors for multi-frame depth estimation. 
	\item A distillation loss based on the minimum reprojection error to filter out the multi-frame depth predictions that may have large errors. 
	\item A new state of the art on the KITTI and Cityscapes datasets in self-supervised monocular depth estimation. The code and models will be available at \href{https://github.com/xjixzz/MISM}{https://github.com/xjixzz/MISM}.
\end{itemize}


\section{RELATED WORKS}

\subsection{Single-Frame Depth Estimation} %Self-Supervised Monocular Depth Estimation
Single-frame depth estimation refers to inferring the corresponding pixel-wise depth from a single image, which is an ill-posed problem because there are an infinite number of possible 3D scenes that can correspond to the same image. Early single-frame depth estimation studies \cite{SaxenaCN05, eigen2014depth} focused on supervised methods, which suffer from collecting ground truth depth. To avoid the heavy work of collecting labels, Garg et al. \cite{garg2016unsupervised} proposed the first self-supervised single-frame depth estimation model supervised by view synthesis loss from rectified stereo image pairs. Zhou et al. \cite{zhou2017unsupervised} extended the self-supervised stereo training into a more general form, i.e., self-supervised monocular training, which jointly estimates depth and poses to form view synthesis loss only using unlabeled monocular videos. Monodepth2 \cite{godard2019digging} further improved the training loss to alleviate the problems caused by occlusions and stationary pixels and provided a strong baseline for the following works including our method. 

Following \cite{godard2019digging}, more powerful or efficient network architectures \cite{guizilini20203d, vadepth} and more effective data augmentation strategies \cite{he2022ra} have been proposed to improve single-frame depth. In addition, other tasks such as flow estimation \cite{ranjan2019competitive}, bird’s-eye-view scene layout estimation \cite{zhao2022jperceiver}, and semantic segmentation \cite{lee2021learning, ma2022towards} were introduced to provide extra information for single-frame depth estimation. Furthermore, some works \cite{ren2022adaptive, Petrovai_2022_CVPR} attempted to use knowledge distillation to improve the results of depth estimation, and we will review these works in Section II-C.

\subsection{Multi-Frame Depth Estimation}
In contrast to single-frame methods, multi-frame depth estimation methods can exploit consecutive multiple frames at test time. Among them, some works e.g. \cite{luo2020consistent} iteratively finetune the pretrained single-frame network at test time for global temporal consistency, which suffers from the running speed. A second group of works e.g. \cite{patil} introduce recurrent networks to exploit temporal information for online depth estimation but are limited by implicitly geometric reasoning.

To explicitly reason about the geometry, deep learning-based multi-view stereo (MVS) methods \cite{yao2018mvsnet, yang2021self, ding2022kd} adopt plane-sweep stereo architectures to build 3D cost volumes from the features of multiple 2D images via differentiable warping. However, these methods focus on simple static scenes and assume that the camera pose is known in advance, which limits their applicability in more complex scenarios. 

To infer the depth from multiple images with unknown poses, Watson et al. \cite{Watson_2021_CVPR} proposed a self-supervised model with the improved multi-view plane-sweep stereo architecture. Manydepth \cite{Watson_2021_CVPR} leverages the estimated single-frame depth to update the minimum and maximum depth values of the whole scene to alleviate the scale ambiguity, and provides supervision for the multi-frame depth network in the region where matching costs do not work. Based on \cite{Watson_2021_CVPR}, \cite{feng2022disentangling} disentangles object motions to overcome the mismatching problem using dynamic category segmentation masks and single-frame depth, and \cite{Guizilini_2022_CVPR} uses an attention-based matching mechanism to improve multi-frame matching for cost volume generation. More recently, \cite{Wang2022CraftingMC} leverages the single-frame depth and the magnitude of the estimated velocity as prior information to determine the search space for multi-frame depth and further uses an additional uncertainty-based network to fuse the single-frame depth and multi-frame depth. Similar to \cite{Wang2022CraftingMC}, \cite{long_two} also utilizes the single-frame depth as prior depth but fuses the single-frame and multi-frame information in a multi-scale manner. However, none of these works exploit the mutual influence between single-frame and multi-frame depth estimation in a comprehensive way as our method.

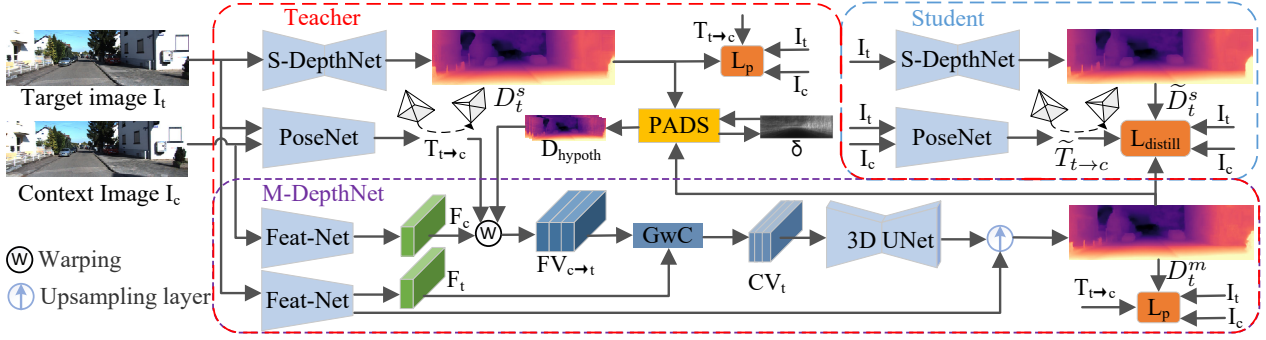
\begin{figure*}[h]
	%\vspace{10pt}
	\centering
	\begin{overpic}[scale=0.9]{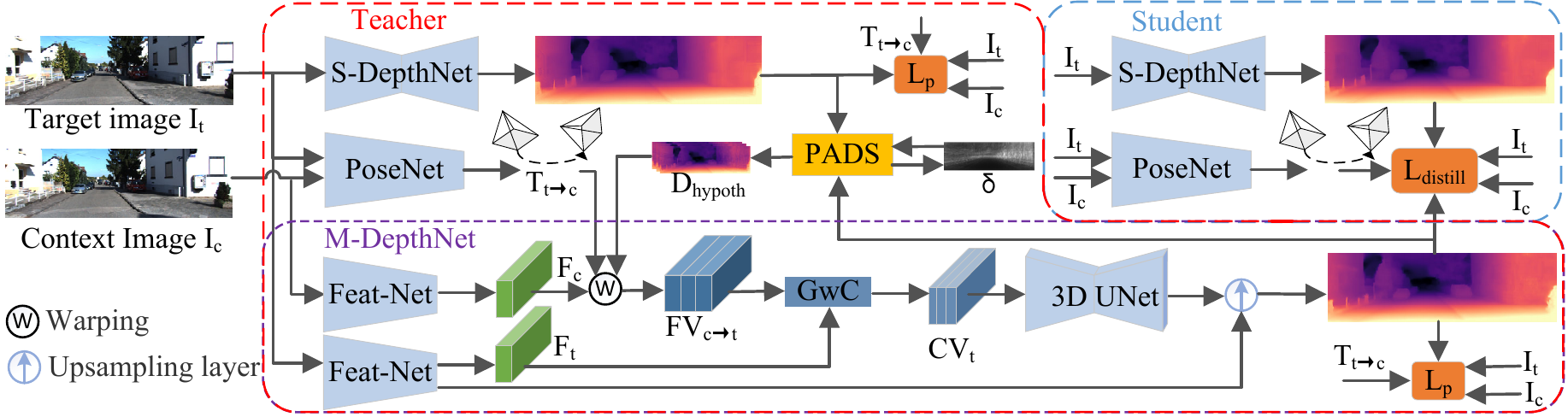}
		\put(39,18.3){$D_{t}^s$}
		\put(92.3,4.5){$D_t^m$}
		\put(83.5,13.5){$\widetilde{T}_{t \rightarrow c}$}
		\put(92.3,18){$\widetilde{D}_t^s$}
	\end{overpic}
	\caption{Overview of our pipeline. First, we train the teacher model consisting of Single-frame Depth Network (S-DepthNet), PoseNet, and Multi-frame Depth Network (M-DepthNet), in which the Pixel-wise Adaptive Depth Sampling (PADS) module generates the hypothesized depth $D_{hypoth}$ for Group-wise Correlation (GwC) based cost volume generation from single-frame depth $D_{t}^s$ and pixel-wise sampling width adjuster $\delta$, and simultaneously updates $\delta$ with $D_t^m$ and $D_t^s$. Then, the trained teacher model generates pseudo labels to guide the student model with the distillation loss. Last, the trained student model replaces the S-DepthNet and PoseNet of the teacher model to help M-DepthNet to produce better results during inference.}
	\label{fig_pipeline}
	%\vspace{-10pt}
\end{figure*}

\subsection{Knowledge Distillation}
Knowledge distillation was originally proposed to compress a large model into a lightweight model without a large performance drop via the teacher-student architecture, which has been applied to many vision tasks \cite{gou2021knowledge}, including MDE \cite{ 9811802}. As for self-supervised depth estimation, research on applying distillation learning to self-supervised MVS methods \cite{yang2021self, ding2022kd} or self-supervised stereo training methods \cite{pilzer2019refine, 9405457, zhou2022self} has been going on for several years.  However, until recently, a few works \cite{ZhouT0M22, ren2022adaptive, Petrovai_2022_CVPR} started to exploit knowledge distillation to improve single-frame depth estimation in the self-supervised monocular setting. Among these methods, \cite{ZhouT0M22} learns two task-dependent uncertainty maps to weight the pseudo label loss and self-supervised photometric loss respectively for more accurate single-frame depth, and \cite{ren2022adaptive} selects the optimal prediction from multiple predictions of the multi-stream ensemble network to help train the student network. Nevertheless, both of these two works ignore temporal information. Closest to our model in spirit is the work of Petrovai et al. \cite{Petrovai_2022_CVPR}, which leverages a self-distillation training strategy to distill the high-resolution pseudo labels with the 3D consistency filtering strategy. Instead of leveraging temporal information to distill pseudo labels via post-processing as \cite{Petrovai_2022_CVPR}, we directly use the multi-frame depth estimation network to generate pseudo labels for training the single-frame depth network and the trained single-frame depth network is further used to boost the performance of multi-frame depth estimation.

\section{METHOD}
In this section, we first introduce the overall pipeline. Then, we introduce the paradigms for self-supervised single-frame and multi-frame depth estimation. After that, we describe the PADS module, which provides effective hypothesized depth for cost volume generation of multi-frame depth estimation. Finally, we introduce the distillation learning to train the student model with the supervision from photometric loss and pseudo labels generated by the teacher model. 

\subsection{Method Overview}
To exploit the mutual influence between single-frame and multi-frame depth estimation, we propose a two-stage training pipeline as shown in Fig. \ref{fig_pipeline}. In the first stage, we train the Multi-frame Depth Network (M-DepthNet) in a self-supervised manner. Following MOVEDepth \cite{Wang2022CraftingMC}, we additionally introduce Single-frame Depth Network (S-DepthNet) and Pose Network (PoseNet) when training M-DepthNet. The output of S-DepthNet is used as the input of the PADS module, which generates the pixel-wise depth range for M-DepthNet. And the output of PoseNet is used to construct cost volumes via warping. Thus, the three networks together form the teacher model and can be jointly optimized. In the second stage, the student model consisting of S-DepthNet and PoseNet is trained by combing the supervision of the teacher model and self-supervision, in which pseudo labels only work when they produce small photometric errors. During inference, the output of the student model is used to guide the cost volume generation for M-DepthNet, which helps improve the accuracy of M-DepthNet.

\subsection{Self-Supervised Single-Frame Depth Estimation}
The objective of self-supervised single-frame depth estimation is to minimize the photometric error between the target image $I_t$ and the synthesized image derived from the predicted target depth map $D_t^s$. As shown in the upper left part of Fig. \ref{fig_pipeline}, self-supervised monocular training jointly optimizes S-DepthNet and PoseNet. S-DepthNet takes $I_t$ as input and outputs the corresponding depth map $D_t^s$, while PoseNet takes $I_t$ and the context image $I_c$ as input and estimates the ego-motion $T_{t \rightarrow c}$. Assuming that the camera intrinsic matrix $K$ is known, then we can project the 2D pixel coordinates $(u, v)$ of $I_t$ to $(u_c, v_c)$ in $I_c$ as follows: 
\begin{equation} \label{pc}
	(u_c, v_c) = KT_{t \rightarrow c}D_t^s(u, v)K^{-1}(u, v, 1)^T,
\end{equation}
where the conversion between homogeneous and inhomogeneous coordinates is omitted for notational simplicity. As in \cite{godard2019digging}, we use bilinear interpolation denoted as $\bigl<\cdot\bigr>$ to sample the context pixel $(u_c, v_c)$ to obtain the synthesized target image:
\begin{equation} \label{Ict}
	I_{c \rightarrow t}(u, v) = I_c\bigl<(u_c, v_c)\bigr>.
\end{equation}
Following \cite{godard2017unsupervised}, we compute the weighted sum of L1 loss and structural similarity (SSIM) to form the photometric loss:
\begin{equation} \label{lpe}
	PE(I_a, I_{b}) = \alpha ||I_a-I_{b}||_1 + (1-\alpha)\frac{1-SSIM(I_a, I_{b})}{2}, 
\end{equation}
where $\alpha=0.15$. To address the occlusion problem, we adopt the per-pixel minimum reprojection loss as in \cite{godard2019digging}, i.e.
\begin{equation} \label{lp}
	L_{p} = \min_{c\in\{t-1, t+1\}} PE(I_t, I_{c \rightarrow t}) . 
\end{equation}
Similar to \cite{godard2019digging}, we also apply the auto-masking strategy to generate the mask $\mu$ for removing the stationary pixels from $L_{p}$. Following \cite{godard2017unsupervised}, we also use the edge-aware smooth loss:
%To enforce disparity smoothness in textureless regions, we follow \cite{godard2017unsupervised} to use the edge-aware smooth loss:
\begin{equation} \label{lsm}
	L_{sm} = |\partial_ud_t^*|e^{-|\partial_uI_t|} + |\partial_vd_t^*|e^{-|\partial_vI_t|} , 
\end{equation}
where $d_t^*$ is the mean-normalized inverse depth. Like \cite{godard2019digging}, we also compute the multi-scale photometric loss when multi-scale depth predictions are available. Thus, the final self-supervised loss for single-frame depth $D_t^s$ is formulated as:
\begin{equation} \label{lbase}
	L_{self}(D_t^s) = \frac{1}{S}\sum_{i=0}^{S-1}\mu L_{p} + \lambda_{sm}L_{sm}, 
\end{equation} 
where $S$ represents the number of multi-scale depth maps and $\lambda_{sm}$ is set to $10^{-3}$ as in \cite{godard2019digging}.

\subsection{Self-Supervised Multi-Frame Depth Estimation}
The diagram of the Multi-frame Depth Network (M-DepthNet) is shown at the bottom of Fig. \ref{fig_pipeline}. As in \cite{Wang2022CraftingMC}, M-DepthNet takes as input two $H \times W \times 3$ images, $I_t$ and $I_c$, and uses a shared Feature Network (Feat-Net) to extract the $h \times w \times C$ features $F_t$ and $F_c$ respectively, where $h=H/4$ and $w=W/4$. Then, similar to (\ref{pc}) and (\ref{Ict}), the context feature $F_c$  is warped into the target view to obtain the feature volume $FV_{c \rightarrow t}$ according to the estimated relative pose $T_{t \rightarrow c}$ and the hypothesized discrete depth candidates $D_{hypoth} \in \mathbb{R}^{N \times h \times w}$, where $N$ is the number of depth candidates for each pixel. Next, group-wise correlation \cite{guo2019group} is applied to construct the cost volume $CV_t \in \mathbb{R}^{N \times G \times h \times w}$, where $G$ is the number of groups that $C$-Channel feature volume $FV_{c \rightarrow t}$ is divided into. Subsequently, a 3D UNet \cite{ronneberger2015u} is used to regularize the cost volume to obtain the probability volume $P_t \in \mathbb{R}^{N \times h \times w}$, and local-max operation \cite{Wang_2022_CVPR} is performed to generate the low-resolution depth map $D_t^{l} \in \mathbb{R}^{h \times w}$  as
\begin{equation} \label{Dl}
	D_t^l(u, v) = \sum_{i=x -r}^{x +r} D_{hypoth}(i, u, v) \frac{P_t(i, u, v)}{\sum_{j=x -r}^{x +r}P_t(j, u, v)}, 
\end{equation}
where $x$ is the index of the maximum value of the 1D vector $P_t(:, u, v)$ and $r$ is the radius of the local window. Finally, the convex upsampling layer \cite{teed2020raft} is used to interpolate the $D_t^{l}$ to output the final multi-frame depth map $D_t^m \in \mathbb{R}^{H \times W}$. We can calculate the self-supervised loss $L_{self}(D_t^m)$ like (\ref{lbase}). Thus, the loss function to jointly optimize all networks of the teacher model is formulated as
\begin{equation} \label{lteacher}
	L_{teacher} = L_{self}(D_t^s) +  L_{self}(D_t^m). 
\end{equation}

\subsection{Pixel-wise Adaptive Depth Sampling}
As described in the previous subsection, generating cost volume requires sampling the depth candidates $D_{hypoth}$. \cite{Watson_2021_CVPR} iteratively updates the depth range $[d_{min}, d_{max}]$ for the whole scene according to the estimated single-view depth during training and fixed the two parameters $d_{min}$ and $d_{max}$ during inference, which is computationally expensive since the learned depth range needs to cover the depths of all viewpoints in the scene. To narrow the depth range for different views, some approaches \cite{Wang2022CraftingMC, long_two} take the estimated single-view depth as the geometric center for depth sampling, and additionally use one or more predefined hyperparameters to determine the width of the depth range. Furthermore, \cite{Wang2022CraftingMC} leverages the magnitude of the velocity estimated by PoseNet to adjust the width of the image-wise depth range but suffers from the challenging task to estimate the absolute scale of the predicted velocity before training. More importantly, all these methods ignore the distribution difference for the width of the sampling range in the pixel space and fail to provide pixel-wise adaptive depth range to build efficient cost volumes.

To better exploit the spatial distribution of scene depth, we propose the PADS module, which adopts a learnable uncertainty map $\delta \in \mathbb{R}^{h \times w}$ to indicate the pixel-wise relative width of the sampling range. All elements in $\delta$ are initialized to one. Following \cite{Wang2022CraftingMC, long_two}, we also use the single-frame depth $D_t^s$ as prior depth to determine the geometric center of the search space for per-pixel depth candidates. Considering the difference in the resolution of the predicted depth map and cost volume, we first downsample $D_t^s$ and $D_t^m$ to obtain $D_t^{s, l}$ and $D_t^{m, l}$ respectively. Let $D_{min}$ and $D_{max}$ denote the minimum and maximum depth map with a resolution of $h \times w$, respectively. Then we can specify the sampling range:
\begin{equation} \label{depth_range}
	\begin{cases}
		D_{min} = D_t^{s, l} / (1 + \delta), \\
		D_{max} = D_t^{s, l}(1 + \delta). 
	\end{cases} 
\end{equation}

When training M-DepthNet, we adopt the exponential moving average strategy to update $\delta$ according to the difference between  $D_t^{s, l}$ and $D_t^{m, l}$:
\begin{equation} \label{EMA}
	\delta \leftarrow 0.99\delta + 0.01 \delta^{\prime}, 
\end{equation} 
\begin{equation} \label{delta_t}
	\delta^{\prime} = \beta (\max(D_t^{s, l} / D_t^{m, l}, D_t^{m, l} / D_t^{s, l}) - 1).
\end{equation}
Here, $\beta$ is a hyperparameter greater than 1 to avoid the estimated multi-frame depth falling on the boundary or even out of the sampling range. In our setting, $\beta$ is set to $1.2$. The learned $\delta$ is visualized as Fig. \ref{vis_ablation}(a), which reflects the estimated uncertainty distribution of single-frame depth for the target scene. Similar to \cite{Watson_2021_CVPR}, we save $\delta$ as part of the model weights after training and keep $\delta$ fixed during inference. 

As in \cite{Wang2022CraftingMC}, according to $D_{min}$ and $D_{max}$ determined by \eqref{depth_range}, we then uniformly sample in the inverse depth space to obtain $D_{hypoth}$, i.e.
\begin{equation} \label{depth_hypoth}
	D_{hypoth}(i) = 1/(\frac{i}{N-1}(\frac{1}{D_{min}} - \frac{1}{D_{max}}) + \frac{1}{D_{max}}),
\end{equation}
where $i = 0, 1, ..., N-1.$ Thus, the depth candidates $D_{hypoth}$ are used to generate the pixel-wise adaptive cost volume for multi-frame depth estimation as described in the previous subsection. Compared to the previous sampling strategies \cite{Watson_2021_CVPR, Wang2022CraftingMC}, the PADS module is capable of adjusting the sampling range at a finer granularity, which helps improve the accuracy of multi-frame depth estimation.

\begin{table*}[t]
	%\vspace{10pt}
	\setlength{\abovecaptionskip}{-5pt}
	\setlength{\belowcaptionskip}{0pt}
	\setlength{\tabcolsep}{3pt}
	\caption{Quantitative results on the Eigen split of KITTI dataset with the raw and improved ground truth}
	\label{table_kitti}	
	\begin{center}
		\begin{threeparttable}
			\begin{tabular}{c|c|c|c|ccc||cccc|ccc}
				\hline
				\multirow{2}{*}{} &\multirow{2}{*}{Method} & \multirow{2}{*}{Train}  & Test &
				\multirow{2}{*}{\#Params.}  &
				\multirow{2}{*}{MACs}  &
				\multirow{2}{*}{Time}  &  
				\multicolumn{4}{c|}{The lower the better} &
				\multicolumn{3}{c}{The higher the better}\\
				\cline{8-14}
				&&&Frames&&&&Abs Rel & Sq Rel & RMSE & RMSE log & $\delta_1$ & $\delta_2$ & $\delta_3$\\
				\hline
				\multirow{15}{*}{\rotatebox{90}{Raw GT}}  
				&Monodepth2 \cite{godard2019digging} 		&M& 1 & 14.3M	&8.0G	&1.4ms	& 0.115 & 0.903 & 4.863 & 0.193 & 0.877 & 0.959 & 0.981 \\
				&PackNet-SfM \cite{guizilini20203d}			&M& 1 & 128.3M	& 205.2G	& 27.4ms & 0.111 & 0.785 & 4.601 & 0.189 & 0.878 & 0.960 & 0.982 \\
				&VADepth \cite{vadepth}  &M& 1 & 18.8M & 9.7G &3.0ms	& 0.104 & 0.774  &   4.552  &  0.181 & 0.892  &   0.965  &   0.983 \\
				&Ma et al. \cite{ma2022towards} 			&M+Sem& 1 & 30.3M & - &-	& 0.104 & 0.690 & 4.473 & 0.179 & 0.886 & 0.965 & 0.984 \\ 
				&SD-SSMDE (ResNet50) \cite{Petrovai_2022_CVPR}			&M& 1 & - & 18.6G &-	& 0.100 & 0.661 & 4.264 & 0.172 & 0.896 & 0.967 & 0.985 \\ 	
				&SUB-Depth \cite{ZhouT0M22} 				&M& 1 & - & - &-	& 0.099 & 0.695 & 4.326 & 0.175 & 0.900 & 0.966 & 0.984 \\
				&RA-Depth \cite{he2022ra} 			&M& 1 & 10.0M & 10.8G &3.4ms	& 0.096 & 0.632 & 4.216 & 0.171 & 0.903 & 0.968 & 0.985 \\
				\cline{2-14}
				&ManyDepth \cite{Watson_2021_CVPR}					&M& 2 (-1, 0) & 26.9M & 15.1G &5.2ms	& 0.098 & 0.770 & 4.459 & 0.176 & 0.900 & 0.965 & 0.983 \\
				&DynamicDepth \cite{feng2022disentangling}				&M+Sem& 2 (-1, 0) & - & - &-	& 0.096 & 0.720 & 4.458 & 0.175 & 0.897 & 0.964 & \underline{0.984} \\
				&Long et al. \cite{long_two}				&M& 2 (-1, 0) & - & 15.6G &-	& 0.097 & 0.731 & 4.392 & 0.176 & 0.901 & 0.965 & 0.983 \\
				&MOVEDepth (ResNet18) \cite{Wang2022CraftingMC}\dag	&M& 2 (-1, 0) & 28.2M & 20.2G &	5.0ms& 0.094  &   0.704  &   4.389  &   0.175  &   0.902  &   0.965  &   0.983  \\ 
				&DepthFormer \cite{Guizilini_2022_CVPR}				&M& 2 (-1, 0) & 28.7M & 174.7G &-	& 0.090 & \underline{0.661} & \underline{4.149} & 0.175 & \underline{0.905} & \underline{0.967} & \underline{0.984} \\
				&MOVEDepth (PackNet) \cite{Wang2022CraftingMC}		&M& 2 (-1, 0) & 142.2M & 217.3G &28.4ms	& \underline{0.089} & 0.663 & 4.216 & \underline{0.169} & 0.904 & 0.966 & \underline{0.984} \\
				\cline{2-14}
				&Ours (ResNet18)				&M& 2 (-1, 0) & 28.2M & 20.2G &4.1ms	&   0.092  &   0.683  &   4.331  &   0.172  &   \underline{0.905}  &   0.966  &   \underline{0.984}  \\
				&\textbf{Ours}		&M& 2 (-1, 0) & 23.9M & 22.9G &6.1ms	&   \textbf{0.086}  &   \textbf{0.613}  &   \textbf{4.096}  &   \textbf{0.165}  &   \textbf{0.915}  &   \textbf{0.969}  &   \textbf{0.985}  \\ 
				\hline\hline
				\multirow{13}{*}{\rotatebox{90}{Improved GT}}
				&Eigen et al. \cite{eigen2014depth} 		&D& 1 &-&-&-	& 0.190 & 1.515 & 7.156 & 0.270 & 0.692 & 0.899 & 0.967 \\
				&DORN \cite{fu2018deep}  		&D& 1 &-&-&-	& 0.072 & 0.307 & 2.727 & 0.120 & 0.932 & 0.984 & \underline{0.994} \\
				&Adabins \cite{Bhat_2021_CVPR} 	&D& 1 &-&-&-	& \underline{0.058} & \underline{0.190} & \underline{2.360} & \underline{0.088} & \underline{0.964} & \underline{0.995} & \textbf{0.999} \\
				
				&NeW CRFs \cite{Yuan_2022_CVPR} 		&D& 1 &-&-&-	& \textbf{0.052} & \textbf{0.155} & \textbf{2.129} & \textbf{0.079} & \textbf{0.974} & \textbf{0.997} & \textbf{0.999} \\	
				\cline{2-14}
				&Monodepth2 \cite{godard2019digging} 		&M& 1 & 14.3M	&8.0G &	-& 0.090 & 0.545 & 3.942 & 0.137 & 0.914 & 0.983 & 0.995 \\
				&PackNet-SfM \cite{guizilini20203d}			&M& 1 & 128.3M	& 205.2G &	-& 0.078 & 0.420 & 3.485 & 0.121 & 0.931 & 0.986 & 0.996 \\
				&RA-Depth \cite{he2022ra}\dag 			&M& 1 & 10.0M & 10.8G &-	&   0.074  &   0.362  &   3.345  &   0.113  &   0.940  &   0.990  &   0.997  \\
				\cline{2-14}
				&Patil et al. \cite{patil}					&M& N & - & 16.9G &-	& 0.087 & 0.495 & 3.775 & 0.133 & 0.917 & 0.983 & 0.995 \\
				
				&ManyDepth \cite{Watson_2021_CVPR}					&M& 2 (-1, 0) & 26.9M & 15.1G &-	& 0.070 & 0.399 & 3.455 & 0.113 & 0.941 & \underline{0.989} & \underline{0.997} \\
				&Long et al. \cite{long_two}				&M& 2 (-1, 0) & - & 15.6G &	-& 0.068 & \underline{0.366} & \underline{3.338} & 0.110 & \underline{0.946} & \underline{0.989} & \underline{0.997} \\
				
				&MOVEDepth (ResNet18) \cite{Wang2022CraftingMC}\dag	&M& 2 (-1, 0) & 28.2M & 20.2G &-	&   0.065  &   0.377  &   3.449  &   0.112  &   0.942  &   0.988  &   0.996  \\
				\cline{2-12}
				&Ours (ResNet18)				&M& 2 (-1, 0) & 28.2M & 20.2G &-	&   \underline{0.064}  &   0.369  &   3.390  &   \underline{0.108}  &   \underline{0.946}  &   0.988  &   0.996  \\
				&\textbf{Ours}		&M& 2 (-1, 0) & 23.9M & 22.9G &	-&   \textbf{0.058}  &   \textbf{0.302}  &   \textbf{3.070}  &   \textbf{0.098}  &   \textbf{0.955}  &   \textbf{0.992}  &   \textbf{0.998}  \\
				\hline
			\end{tabular}
			\begin{tablenotes}
				\footnotesize
				\item  All self-supervised methods are tested with the resolution of $192\times640$. ``$\dag$" means evaluation on the pretrained models from github. The best scores for each subsection are in \textbf{bold} and the second are \underline{underlined}. In the ``Train" column, we list the training data for each method with D --- ground truth Depth, M --- unlabeled Monocular videos, Sem --- Semantic labels. In the ``Test Frames" column, ``N" refers to taking a long sequence of frames as input to predict the target depth map. In the ``Time'' column, we list the inference time to generate one depth map by averaging the inference time of all 697 test images with a batch size of 16.
			\end{tablenotes}
		\end{threeparttable}
	\end{center}
	\vspace{-15pt}
\end{table*}

\subsection{Distillation Learning} 
Considering the performance gap between single-frame and multi-frame depth networks, we further transfer the knowledge from M-DepthNet to S-DepthNet. As shown in Fig. \ref{fig_pipeline}, the student model is composed of S-DepthNet and PoseNet, and the outputs of S-DepthNet and PoseNet are $\widetilde{D}_t^s$ and $\widetilde{T}_{t \rightarrow c}$ respectively. The teacher model generates pseudo labels $D_t^m$  for supervising the student model.  

Since the teacher network might produce results with large errors for some pixels, it is necessary to filter out the pixels with large errors. Inspired by \cite{godard2019digging}, we introduce the minimum reprojection error to construct distillation loss for filtering out the multi-frame depth values that generate larger errors than the student single-frame depth. Given $\widetilde{T}_{t \rightarrow c}$, we can synthesize the images $\widetilde{I}_{c \rightarrow t}^s$ and $\widetilde{I}_{c \rightarrow t}^m$ according to $\widetilde{D}_t^s$ and $D_t^m$ respectively. Then, we can compare their photometric errors and generate the mask:
\begin{equation} \label{Mask}
	M = \left[PE(I_t, \widetilde{I}_{c \rightarrow t}^m) < PE(I_t, \widetilde{I}_{c \rightarrow t}^s)\right], 
\end{equation} 
where $[\cdot]$ is the Iverson bracket.
Following \cite{eigen2014depth}, we adopt the scale-invariant error between $D_t^m$ and $\widetilde{D}_t^s$ as the pseudo-label based regression loss:
\begin{equation} \label{L_si}
	L_{si} = \sqrt{\frac{1}{n} \sum_{u, v} (d(u,v))^2 - \frac{\gamma}{n^2} \big(\sum_{u, v} d(u,v)\big)^2},
\end{equation} 
where $d(u,v)=(log(\widetilde{D}_t^s(u,v)) - log(D_t^m(u,v)))M(u, v)$, $n$ represents the number of elements with a value of 1 in $M$, and $\gamma =1.0$. To provide supervision for the pixels where
pseudo labels do not work, we also introduce the self-supervised loss ${L}_{self}(\widetilde{D}_t^s)$ as in (\ref{lbase}) to construct the final distillation loss:
\begin{equation} \label{L_distill}
	L_{distill} = {L}_{self}(\widetilde{D}_t^s) + \lambda_{si}L_{si},
\end{equation}
where $\lambda_{si} = 0.1$. In this way, the trained student model produces more accurate single-view depth than the S-DepthNet of the teacher model, thanks to the distilled geometric matching knowledge from the multi-frame network. Ultimately, we use the distilled student model to guide the M-DepthNet to obtain better depth estimation at test time.

\begin{figure*}[t]
	\centering
	%\vspace{10pt}
	\setlength{\abovecaptionskip}{-2pt}
	%\begin{overpic}[scale=0.9, grid, tics=5]{./figs/VAB.pdf}
	\begin{overpic}[scale=0.125]{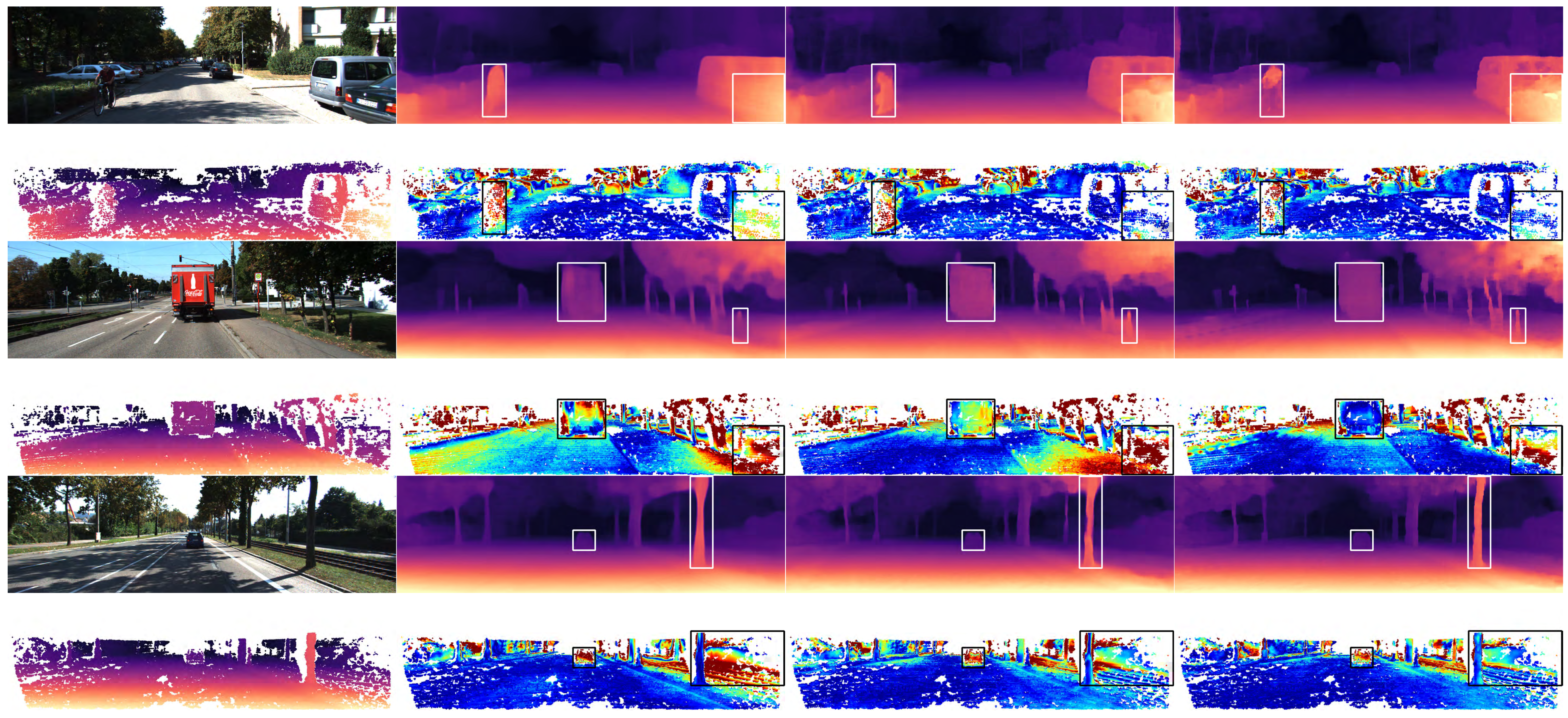}
		\footnotesize
		\put(8,46){Input \& GT}
		\put(33,46){Manydepth \cite{Watson_2021_CVPR}}
		\put(56,46){MOVEDepth \cite{Wang2022CraftingMC}}
		\put(85,46){Ours}
	\end{overpic}
	\caption{Qualitative results on the Eigen test split of KITTI dataset. Rows 2, 4, 6 provide the equivalently colormapped error maps for the metric Abs. Rel. relative to the improved depth \cite{uhrig2017sparsity}, from small (blue) to large (red) errors. The GT depth is interpolated for better visualization. Compared to other methods \cite{Watson_2021_CVPR, Wang2022CraftingMC}, our model not only preserves better details for various objects but also predicts depth maps with small errors. White and black boxes highlight the difference for the predicted depth and error maps, respectively. Best viewed in color and zoom in.}
	\label{fig_kitti}
	\vspace{-15pt}	
\end{figure*}
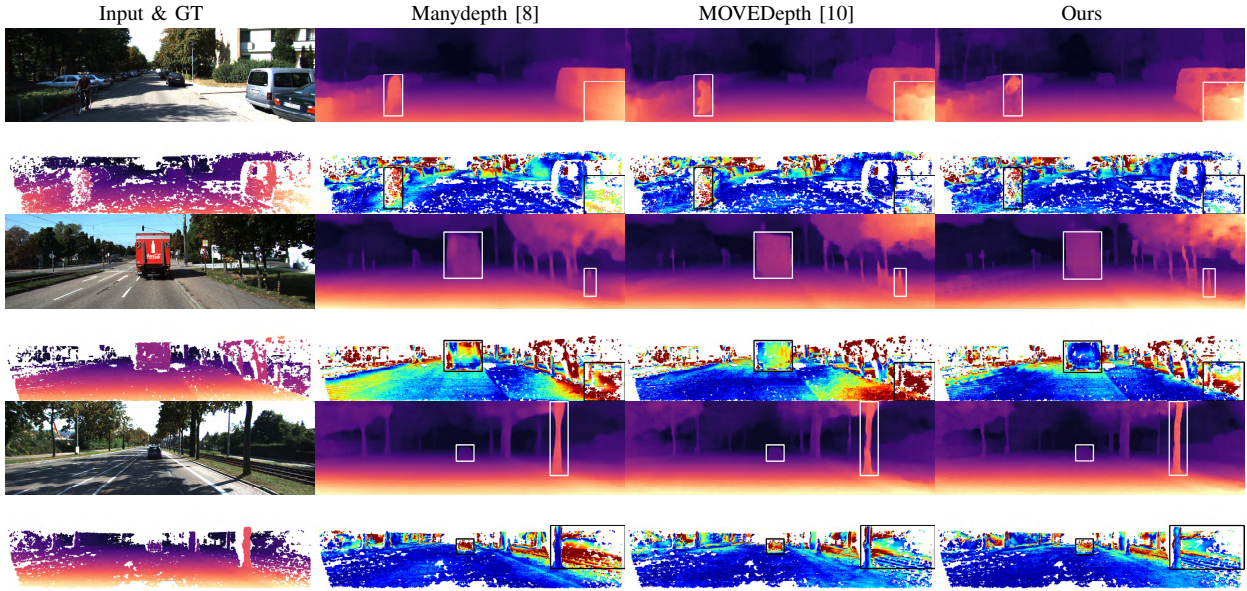

\section{EXPERIMENTAL RESULTS}
\subsection{Datasets}
We conduct experiments on the KITTI \cite{geiger2012we} and Cityscapes \cite{cordts2016cityscapes} datasets to verify the effectiveness of our method with the metrics proposed in \cite{eigen2014depth}. KITTI is one of the most widely-used datasets for depth estimation, which covers various outdoor scenes. We follow \cite{zhou2017unsupervised} to adopt the Eigen split \cite{eigen2014depth} and remove the static frames, which results in 39810/4424/697 training, validation, and test images. We also evaluate our model on the improved depth maps from \cite{uhrig2017sparsity}, containing 652 test images. As for Cityscapes \cite{cordts2016cityscapes}, it is a large dataset that is comprised of video sequences captured in streets from 50 cities. Following \cite{zhou2017unsupervised}, we train on the 69731 monocular triplets and evaluate on the 1525 test images. As in \cite{zhou2017unsupervised}, the maximum depth of evaluation on both datasets is restricted to 80m.  

\subsection{Implementation Details}
In our experiments, we adopt the HRNet \cite{wang2020deep} based architecture in \cite{he2022ra} as S-DepthNet unless otherwise stated, where the number of output scales $S$ is set to 1. The PoseNet is a modified ResNet18 \cite{he2016deep} as in \cite{godard2019digging}. Both the backbones of S-DepthNet and PoseNet are initialized with weights pretrained on ImageNet \cite{deng2009imagenet}. Following \cite{Wang2022CraftingMC}, we adopt a four-stage FPN \cite{lin2017feature} as the Feat-Net, where the number of feature channels $C=32$. As for cost volume generation, both the number of depth candidates $N$ and the number of groups $G$ are set to 16. Similar to \cite{Watson_2021_CVPR}, we set the input resolution as $192\times640$ for KITTI and $128\times 416$ for Cityscapes. Following \cite{godard2019digging}, we use random color-jitter and flip for data augmentations, and further apply the random image mask strategy when training M-DepthNet as in \cite{Wang2022CraftingMC}. All experiments are performed on a single Nvidia RTX 3090 GPU. Our models are implemented in Pytorch and the batch size is set as 12 and 16 for training and test respectively. Both the teacher model and student model are trained with Adam optimizer for $E$ epochs. $E$ is set to 20 for KITTI and 5 for Cityscapes. The initial learning rate is set to $2\times 10^{-4}$ for the teacher model and $1\times 10^{-4}$ for the student model, dropping by a factor of 10 after $Q$ epochs. $Q$ is 15 for KITTI and 1 for Cityscapes. For KITTI, it takes approximately 13 and 10 hours to train the teacher and student models, respectively. For Cityscapes, it takes about 5 and 4 hours respectively.

\begin{table}[t]
	%\vspace{10pt}
	\setlength{\abovecaptionskip}{-5pt}
	\setlength{\belowcaptionskip}{0pt}
	\caption{Quantitative results on the Cityscapes dataset}
	\label{table_cs}	
	\begin{center}
		\begin{threeparttable}
			\begin{tabular}{c|cccc}
				\hline
				Method&AbsRel$\downarrow$ & SqRel$\downarrow$ & RMSE$\downarrow$ & RMSE log$\downarrow$ \\
				\hline 
				Monodepth2 \cite{godard2019digging} & 0.129 & 1.569 & 6.876 & 0.187 \\
				
				InstaDM \cite{lee2021learning}*			& 0.111 & 1.158 & 6.437 & 0.182 \\
				\hline
				ManyDepth \cite{Watson_2021_CVPR}			& 0.114 & 1.193 & 6.223 &0.170 \\
				
				Long et al.	\cite{long_two}	& 0.113 & 1.093 & 6.119 & 0.170 \\
				
				DynamicDepth \cite{feng2022disentangling}*		& \underline{0.103} & \underline{1.000} & \underline{5.867} & \underline{0.157} \\
				
				\textbf{Ours}	&   \textbf{0.102}  &   \textbf{0.948}  &   \textbf{5.788}  &   \textbf{0.154} \\
				\hline
			\end{tabular}
			\begin{tablenotes}
				\footnotesize
				\item All methods are tested with the resolution of $128\times416$, except InstaDM \cite{lee2021learning} with a resolution of $256 \times 832$. ``*" means that the method requires semantic labels for training. 
			\end{tablenotes}
		\end{threeparttable}
	\end{center}
	\vspace{-15pt}	
\end{table}

\subsection{Depth Evaluation}
To evaluate the performance of our method, we first conduct a comparison of its performance relative to the state-of-the-art self-supervised MDE methods on KITTI with the raw \cite{geiger2012we} and improved ground truth \cite{uhrig2017sparsity}. As shown in Table \ref{table_kitti}, our method establishes a new state of the art in self-supervised MDE, with competitive model complexity and inference speed.Compared with the best-performing single-frame method \cite{he2022ra} and distillation learning based methods \cite{Petrovai_2022_CVPR, ZhouT0M22}, our model improves the performance by more than $10.4\%$ (on Abs. Rel. with the raw GT). Note that our method only adopts the same S-DepthNet as RA-Depth, but does not adopt the data augmentation strategy and the cross-scale depth consistency loss proposed in \cite{he2022ra}. The multi-frame depth estimation methods that perform closest to our method are DepthFormer \cite{Guizilini_2022_CVPR} and MOVEDepth \cite{Wang2022CraftingMC}. Although neither using the computationally expensive transformer architecture \cite{Guizilini_2022_CVPR} nor adopting the velocity-guided depth sampling strategy and additional depth fusing network \cite{Wang2022CraftingMC}, our model still outperforms these methods in all metrics. For a fair comparison with \cite{Wang2022CraftingMC}, we also list the results using the same ResNet18-based architecture (number of output scales $S=4$) for our method and \cite{Wang2022CraftingMC}, where our method still performs better. Furthermore, the bottom half of Table \ref{table_kitti} shows that our method even compares favorably to some single-frame supervised methods \cite{fu2018deep, Bhat_2021_CVPR}, and narrows the performance gap between self-supervised monocular training methods and the best-performing supervised approach \cite{Yuan_2022_CVPR}.

In addition, we also present the qualitative results in Fig. \ref{fig_kitti}, where our method better preserves the shape of objects and outputs depth maps with smaller errors. 

Moreover, we also compare the results with the current state-of-the-art methods \cite{Watson_2021_CVPR, long_two, feng2022disentangling} on Cityscapes. As shown in Table \ref{table_cs}, our method performs best again, even though DynamicDepth \cite{feng2022disentangling} leverages semantic labels.

\begin{table}[t]
	%\vspace{10pt}
	\setlength{\abovecaptionskip}{-5pt}
	\setlength{\belowcaptionskip}{0pt}
	\caption{Generalization performance on the Cityscapes dataset}
	\label{table_generalization}	
	\begin{center}
		\begin{threeparttable}
			\begin{tabular}{c|cccc}
				\hline
				Method&AbsRel$\downarrow$ & SqRel$\downarrow$ & RMSE$\downarrow$ & RMSE log$\downarrow$\\
				\hline
				ManyDepth \cite{Watson_2021_CVPR}					&   0.170  &   1.789  &   8.357  &   0.236 \\
				
				MOVEDepth \cite{Wang2022CraftingMC}\dag					&   0.164  &   1.780  &   8.678  &   0.238  \\
				\textbf{Ours}	& \textbf{0.150}  &   \textbf{1.492}  &   \textbf{7.810}  &   \textbf{0.216} \\
				\hline
			\end{tabular}
			\begin{tablenotes}
				\footnotesize
				\item ``$\dag$" means evaluation on the pretrained models from github. 
			\end{tablenotes}
		\end{threeparttable}
	\end{center}
	\vspace{-15pt}	
\end{table}   

\subsection{Generalization Performance}
To evaluate the generalization capability across datasets, the model trained on the  KITTI dataset is used to test on the Cityscapes dataset without finetuning. Table \ref{table_generalization} compares the generalization performance with the current self-supervised multi-frame depth estimation methods \cite{Watson_2021_CVPR, Wang2022CraftingMC}, from which we can see that our method achieves better results. These data demonstrate that digging into the complementary information between single-frame and multi-frame may contribute to improving the generalization capability across datasets.

\begin{table}[t]
	%\vspace{10pt}
	\setlength{\abovecaptionskip}{-5pt}
	\setlength{\belowcaptionskip}{0pt}
	\setlength{\tabcolsep}{2pt}
	\caption{Ablation study on KITTI Eigen split for multi-frame depth}
	\label{tab_ablation_m}
	\begin{center}
		\begin{threeparttable}
			\begin{tabular}{c|c|c|cccc|cc}
				\hline
				\multirow{2}{*}{PADS} & \multirow{2}{*}{Distill}	& Min. &\multicolumn{4}{c}{The lower the better} &\multicolumn{2}{|c}{GPU (GB)}\\
				\cline{4-7}\cline{8-9}
				&&Reproj.&AbsRel & SqRel & RMSE & R log& train* & test\\ 
				\hline
				\multicolumn{3}{c|}{Baseline (HRNet18)} &   0.090  &   0.704  &   4.293  &   0.171  & 15.1 & 4.2\\
				\hline
				$\checkmark$ &&&   0.088  &   0.673  &   4.257  &   0.169  & 15.1 & 4.2\\
				&$\checkmark$ &&   0.088  &   0.640  &   4.196  &   0.168  & 15.1/10.4 & 4.2\\
				&  $\checkmark$ &$\checkmark$ &   0.088  &   0.655  &   4.183  &   0.168 & 15.1/10.5 & 4.2 \\
				$\checkmark$ &$\checkmark$ &  &   0.087  &   0.637  &   4.133  &   0.166 & 15.1/10.4 & 4.2 \\
				$\checkmark$ &$\checkmark$ &$\checkmark$  &   \textbf{0.086}  &   \textbf{0.613}  &   \textbf{4.096}  &   \textbf{0.165} & 15.1/10.5 & 4.2  \\
				\hline\hline
				
				\multicolumn{3}{c|}{Baseline (ResNet18)} &  0.096  &   0.760  &   4.499  &   0.178 & 14.7 & 4.0 \\
				\hline
				$\checkmark$ &&&   0.094  &   0.748  &   4.457  &   0.176  & 14.7 & 4.0 \\
				&$\checkmark$ &&   0.094  &   0.714  &   4.379  &   0.175  & 14.7/10.2 & 4.0\\
				&  $\checkmark$ &$\checkmark$ &   0.093  &   0.684  &   4.335  &   0.173 & 14.7/10.3 & 4.0 \\
				$\checkmark$ &$\checkmark$ &  &   \textbf{0.092}  &   0.691  &   4.353  &   \textbf{0.172}& 14.7/10.2 & 4.0  \\
				$\checkmark$ &$\checkmark$ &$\checkmark$  &   \textbf{0.092}  &   \textbf{0.683}  &   \textbf{4.331}  &   \textbf{0.172}  & 14.7/10.3 & 4.0\\
				\hline				
			\end{tabular}
			\begin{tablenotes}
				\footnotesize
				\item ``*'': separate GPU memory required for the two stages of training.
			\end{tablenotes}
		\end{threeparttable}			
	\end{center}
	\vspace{-15pt}
\end{table}

\begin{table}[t]
	%\vspace{10pt}
	\setlength{\abovecaptionskip}{-5pt}
	\setlength{\belowcaptionskip}{0pt}
	\caption{Ablation study on KITTI Eigen split for single-frame depth}
	\label{tab_ablation_s}
	\begin{center}
		\begin{tabular}{c|c|cccc|c}
			\hline
			\multirow{2}{*}{Distill}	& Min. &\multicolumn{4}{c|}{The lower the better} &\multirow{2}{*}{$\delta_1\uparrow$}\\
			\cline{3-6}
			&Reproj.&AbsRel & SqRel & RMSE & R log&\\
			\hline
			\multicolumn{2}{c|}{Baseline (HRNet18)} &   0.102  &   0.741  &   4.470  &   0.179  &   0.896  \\
			\hline
			
			$\checkmark$ &&   0.100  &   0.702  &   4.328  &   0.175  &   \textbf{0.900}  \\
			
			$\checkmark$ &	$\checkmark$ &\textbf{0.099}  &   \textbf{0.676}  &   \textbf{4.287}  &   \textbf{0.174}  &   \textbf{0.900}  \\
			\hline\hline
			\multicolumn{2}{c|}{Baseline (ResNet18)} &   0.115  &   0.885  &   4.799  &   0.192  &   0.876  \\
			\hline
			
			$\checkmark$ &&   0.111  &   0.799  &   4.634  &   \textbf{0.185}  &   0.880  \\
			
			$\checkmark$ &	$\checkmark$ &\textbf{0.110}  &   \textbf{0.786}  &   \textbf{4.603}  &   \textbf{0.185}  &   \textbf{0.884}  \\
			\hline
			\multicolumn{2}{c|}{SUB-Depth \cite{ZhouT0M22}} &   \textbf{0.110}  &   0.821  &   4.648  &   \textbf{0.185}  &   \textbf{0.884}  \\
			\hline				
		\end{tabular}			
	\end{center}
	\vspace{-15pt}
\end{table}

\begin{figure}[t]
	\centering
	%\vspace{10pt}
	%\setlength{\abovecaptionskip}{0pt}
	%\begin{overpic}[scale=0.9, grid, tics=5]{./figs/VAB.pdf}
	\begin{overpic}[scale=0.12]{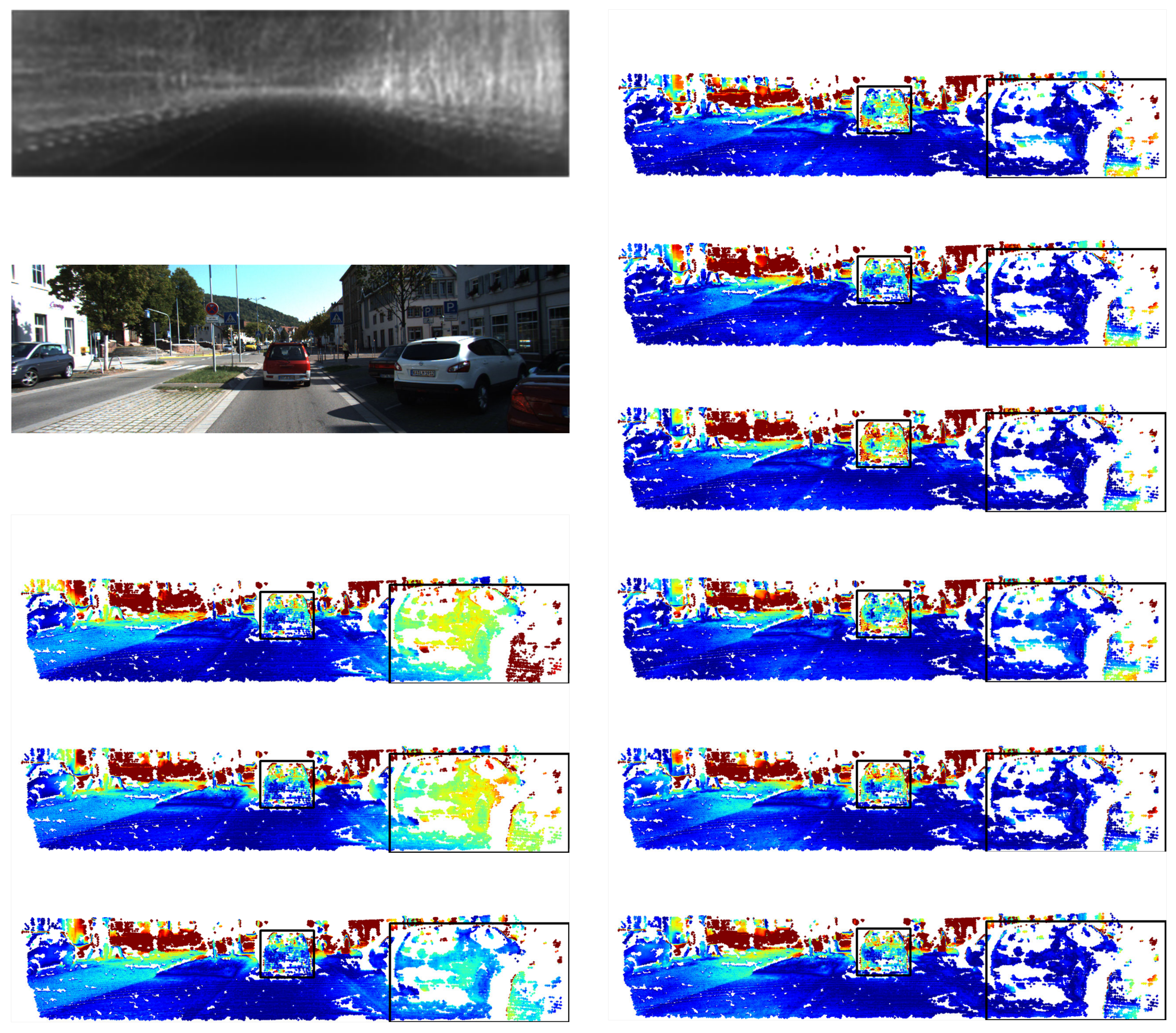}
		\footnotesize
		\put(1,70){(a) The learned uncertainty map $\delta$}  %(higher is brigher)}
	\put(12,48){(b) Target image}
	
	\put(6, -2){(c) Error map of S-Depth}
	\put(2, 40){Baseline}
	\put(2, 25.5){Distill w/o Min. Reproj.}
	\put(2, 11){Distill with Min. Reproj.}
	
	\put(57, -2){(d) error map of M-Depth}
	\put(52.5, 82){Baseline}
	\put(52.5, 68){PADS}
	\put(52.5, 54){Distill w/o Min. Reproj.}
	\put(52.5, 40){Distill with Min. Reproj.}
	\put(52.5, 25.5){PADS + Distill w/o Min. Reproj.}
	\put(52.5, 11){PADS + Distill with Min. Reproj.}
\end{overpic}
\caption{Visualization results of ablation study. The learned $\delta$ is visualized in (a), where high uncertainty corresponding to large sampling range is white, otherwise black. The error maps of single-frame depth and multi-frame depth for different settings are visualized as (c) and (d) respectively.}
\label{vis_ablation}
\vspace{-15pt}	
\end{figure}
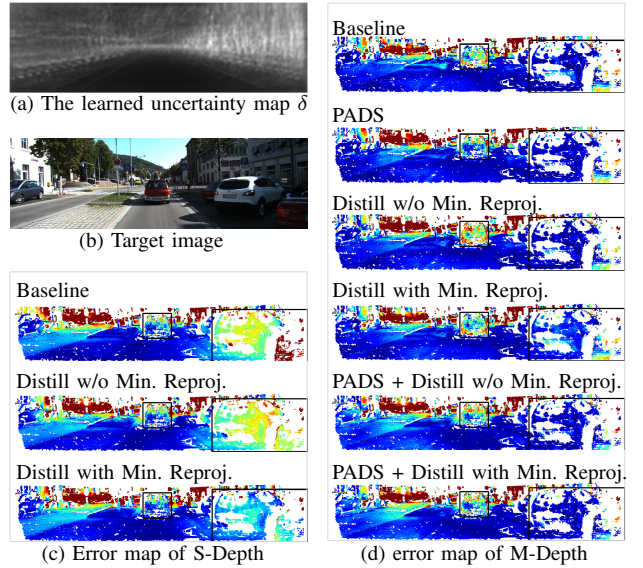

\subsection{Ablation Study}
To understand how much each component of our method contributes to the overall performance of multi-frame depth estimation, we conduct the ablation experiments on KITTI with the raw GT depth. In our experiments, the baseline model is the trained teacher model without distillation learning, in which single-frame depth is set as the geometric center of the sampling range and all elements of $\delta$ are fixed as 0.3 following \cite{Wang2022CraftingMC}. The top half of Table \ref{tab_ablation_m} shows that using the PADS module leads to better scores in all metrics, which proves the effectiveness of the PADS module. Adopting the PADS module does not change the computation complexity but only introduces 7.68K extra parameters of $\delta$, which is negligible compared to the 23.9M parameters of the baseline model. Thus, using the PADS module requires similar GPU memory as the baseline. Adopting distillation learning can achieve greater performance gains than only introducing the PADS module, which reveals the effect of the proposed two-stage training pipeline. Combining the distillation learning loss with the masking strategy based on minimum reprojection error further brings an improvement in the overall performance at the cost of a little more GPU memory (10.4 v.s. 10.5) required for training the student model, which reflects the necessity to mask out the false labels generated by the teacher model. Taking all three components together leads to the most accurate depth predictions and disabling any component may result in performance degradation. These results suggest that all these components are compatible with each other and lead to more utilization of the mutual influence between single-frame and multi-frame depth estimation. In addition, to verify the compatibility of our method with different single-frame depth network architectures, we adopt the ResNet-based architecture in \cite{godard2019digging} for the ablation study. From the bottom half of Table \ref{tab_ablation_m}, we can observe consistent results and draw the same conclusion.

Moreover, we perform an ablation study on the single-frame depth to further analyze the effectiveness of the minimum reprojection based distillation learning. As listed in Table \ref{tab_ablation_s}, the minimum reprojection based distillation learning brings a considerable performance gain for S-DepthNet.  When applying the same backbone (ResNet18), our method performs better than the self-distillation method \cite{ZhouT0M22}, which demonstrates that distillation learning from multi-frame depth estimation is more effective than self-distillation. Considering the performance gap between the distilled S-DepthNet and the M-DepthNet (even the baseline), we further use the trained student model to guide the M-DepthNet to output the final depth map. The consistent visualization results of the HRNet18-based models corresponding to the settings of Table \ref{tab_ablation_m} and Table \ref{tab_ablation_s} are shown in Fig. \ref{vis_ablation}. Taking all results of the ablation study together, we find that both single-frame depth estimation and multi-frame depth estimation do help improve each other.

\section{CONCLUSION}

In this paper, we presented a distillation learning pipeline for self-supervised MDE so that single-frame and multi-frame depth networks can benefit from each other. Thanks to the proposed PADS module and minimum reprojection based distillation loss, our model achieves state-of-the-art performance and generalizes better than the previous methods. 

However, the performance of our method on Cityscapes is still limited, which may suffer from more moving objects compared to KITTI which captures more stationary dynamic objects. In addition, the generalization performance of our method is also unsatisfactory. Note that dynamic scenes or new scenes and cameras not only directly affect M-DepthNet, but also indirectly affect M-DepthNet by affecting PoseNet. Thus, further combining S-DepthNet and M-DepthNet to adapt to highly dynamic scenes or new scenes and cameras might be worthwhile, especially taking PoseNet into account.

% or


%\appendices
%% you can choose not to have a title for an appendix
%% if you want by leaving the argument blank
%\section{}
%% use section* for acknowledgment
%\section*{Acknowledgment}

\ifCLASSOPTIONcaptionsoff
  \newpage
\fi

\end{document}